\documentclass[10pt,conference]{IEEEtran}
\IEEEoverridecommandlockouts

\usepackage{cite}
\usepackage[utf8]{inputenc}
\usepackage{amsmath,amssymb,amsfonts}
\usepackage{algorithm}
\usepackage{algorithmicx}
\usepackage{algpseudocode}
\usepackage{graphicx}
\usepackage{textcomp}
\usepackage{xcolor}
\usepackage{balance}
\usepackage{makecell}
\usepackage{pifont}
\usepackage{url}
\usepackage{booktabs}
\usepackage{tabularx}
\usepackage{enumitem}
\usepackage{hyperref} 
\DeclareRobustCommand{\myurl}[1]{\url{#1}}

\def\BibTeX{{\rm B\kern-.05em{\sc i\kern-.025em b}\kern-.08em T\kern-.1667em\lower.7ex\hbox{E}\kern-.125emX}}

\newcommand{\todo}[1]{\textsf{\textbf{\textcolor{Orange}{[#1]}}}}

\newcommand{\boldpar}[1]{\smallbreak\noindent\textbf{#1.}}

\newcommand{\semcom}{SemCom}

\begin{document}
\title{

\huge{When Robots Exchange Meaning: A Demo of Goal-Oriented Semantic Communications for Collaborative Robotics}}

\author{\IEEEauthorblockN{
Peizheng Li, Xinyi Lin, Sajida Gufran, Adnan Aijaz
}\\ 
\IEEEauthorblockA{
Bristol Research and Innovation Laboratory, Toshiba Europe Ltd., U.K.\\
Email: 
{\{peizheng.li, xinyi.lin, sajida.gufran, adnan.aijaz\}@toshiba-bril.com}
}
}

\maketitle

\begin{abstract}
Collaborative robotics is a representative task-oriented 6G use-case, where communication quality should be reflected in mission execution, environment understanding, and closed-loop operation rather than packet delivery alone. This demo paper presents a robot-edge semantic communication (SemCom) testbed integrating robot-side visual compression, edge-side semantic mapping, and dashboard-based mission interaction. A mobile robot equipped with RGB-D sensing and LiDAR runs ROS~2, while a Jetson Orin edge node performs reconstruction, RTAB-Map mapping, semantic object handling, and browser-based visualization. As an initial proof of concept, RGB frames are encoded on the robot into VQ-VAE tokens using an ONNX Runtime encoder and reconstructed on the edge using a PyTorch decoder. A $320 \times 240$ image is represented by an $80 \times 60$ token grid with a packed payload of 5400 bytes, corresponding to a $42.67\times$ reduction relative to model-input RGB bytes. The reconstructed visual stream is further associated with depth, pose, and 3D mapping information to generate a semantic map for downstream robotic applications. The demo exposes the full path from semantic visual transport to object-level map interaction, and provides a practical platform for future task-aware 6G networking studies at the intersection of \semcom, embodied AI, and physical AI-enabled robotics. A video of the demo is available at \url{https://tinyurl.com/Tos09}.
\end{abstract}

\begin{IEEEkeywords}
Semantic communications, 6G, collaborative robotics, task-oriented networking, VQ-VAE, edge intelligence, O-RAN, robot testbed.
\end{IEEEkeywords}

\section{Introduction}
\label{sec:introduction}
Future 6G systems are expected to evolve from packet-centric optimization towards task-oriented services, where network utility is measured by application outcomes rather than transport metrics alone. Collaborative robotics is a strong example because communication affects whether robots can perceive the environment, update maps, coordinate actions, and complete missions safely.

Semantic communication (\semcom) is therefore attractive for robotics~\cite{strinati2024goal,li2025task, li2025rethinking}, due to its meaning delivery and data compression capabilities. A robot continuously produces high-rate perceptions such as camera, depth, LiDAR, odometry, and state streams, yet only part of this information is directly useful for a given downstream task. Learned semantic representations can reduce communication load while preserving the task-relevant representation needed for monitoring, mapping, and object-level interaction.

This demo paper presents a work-in-progress (WIP) robot-edge testbed designed for RAN integration that organizes the system around three visible functions: semantic compression on the robot, semantic mapping on the edge, and dashboard-based integration for mission control. The underlying communication is provided by a private 5G/Open RAN infrastructure. The contribution is not a fully optimized benchmark, but an end-to-end platform that shows how these vertical functions for semantic processing can be delivered over a networking infrastructure.

\section{Demo System Overview}

\begin{figure}[t]
    \centering
    \includegraphics[width=\linewidth]{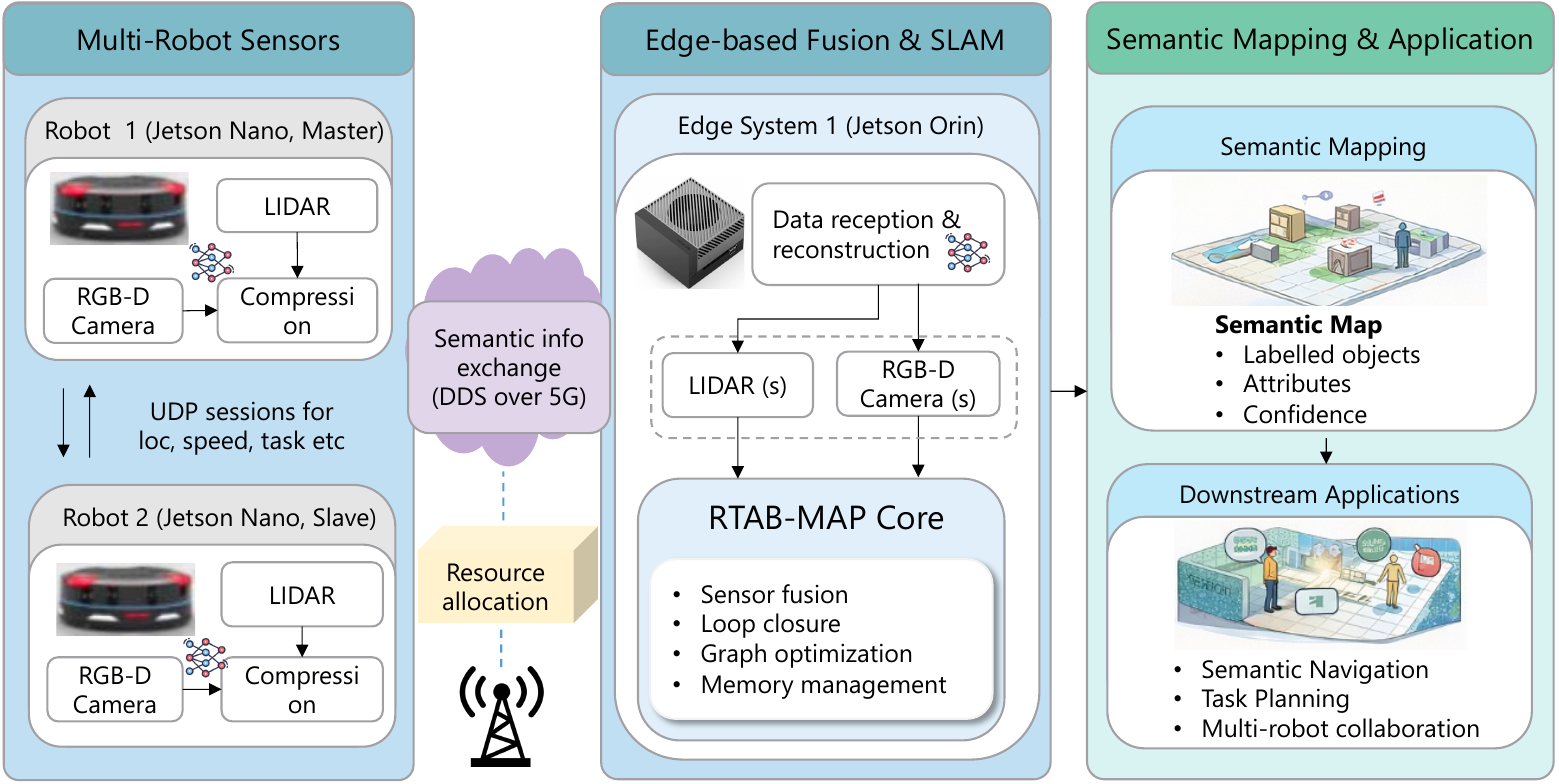}
    \caption{Robot-edge architecture for the collaborative robotics demo.}
    \label{fig:architecture}
\end{figure}

Fig.~\ref{fig:architecture} illustrates the overall robot-edge architecture of the demo system. Multiple mobile robots collect RGB-D camera and LiDAR data, perform local compression, and exchange semantic and control information with the edge over the communication infrastructure. At the edge, a Jetson Orin receives and reconstructs the sensing streams, then performs sensor fusion and simultaneous localization and mapping (SLAM) using the real-time appearance-based mapping (RTAB-Map~\cite{rtabmap_roswiki}) core. The resulting geometric map is further enriched by the semantic mapping module, which maintains labelled objects, attributes, and confidence information. These semantic map outputs support downstream robotic applications such as semantic navigation, task planning, and multi-robot collaboration.

Robot-edge connectivity currently carries RGB-D, LiDAR, odometry, Transform (TF), semantic token messages, and dashboard control information. The wider deployment target is a private 5G Open RAN setup, where the robot acts as a UE and the edge node acts as a multi-access edge computing (MEC) endpoint \cite{aijaz2023open}. This makes the testbed suitable for future studies on how semantic streams and mission feedback can be exposed to semantic and task-aware network control, shaping the resource allocation and scheduling of the communication layer.

\boldpar{Demo workflow}
During a typical run, the robot publishes RGB-D, LiDAR, odometry, and TF streams over ROS 2 while the semantic encoder produces image tokens in parallel. The edge node receives both conventional robotic telemetry and semantic tokens, reconstructs the RGB frames, updates the RTAB-Map representation, and exposes intermediate outputs to the dashboard. The operator can then monitor scene evolution, compare raw and reconstructed views, inspect map growth, and issue high-level commands such as start/stop, pause/resume, and click-to-go navigation. This is useful for a demo setting because the communication, mapping, and mission layers remain visible within one end-to-end loop.
Meanwhile, multi-robot RTAB-MAP perception fusion, coordination and formation control have also been implemented in the system, which can be observed from the attached video.
In the next section, we highlight the main demo features and task-oriented observables (Table~\ref{tab:task_observables}) from the angles of semantic compression, semantic mapping, and dashboard integration.

\section{Demo Highlights}

\begin{figure}[t]
    \centering
    \includegraphics[width=\linewidth]{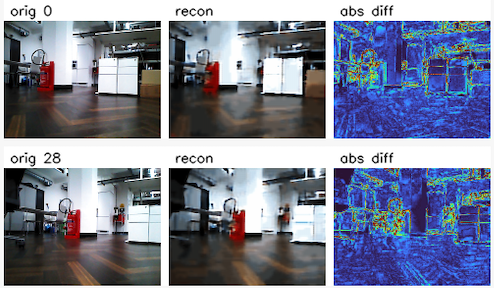}
    \caption{Example original and reconstructed frames from the VQ-VAE pipeline.}
    \label{fig:vqvae_results}
\end{figure}

\boldpar{Semantic compression}
The visual semantic compression and reconstruction pipeline uses a vector-quantized variational autoencoder (VQ-VAE) model~\cite{van2017neural}. Each $320 \times 240$ RGB frame is encoded on the robot into an $80 \times 60$ latent grid, corresponding to 4800 discrete tokens. With a codebook size of 512, each token requires 9 bits and the packed payload is 5400 bytes per frame, compared with 230400 bytes for the model-input RGB frame. This yields a $42.67\times$ model-input-level reduction. The encoder runs with ONNX Runtime on the robot, while the edge decodes the token stream with PyTorch. Offline tests on saved samples achieve 21.7--21.9 dB PSNR, which is sufficient for coarse scene awareness and proof-of-concept semantic transport. Example reconstructions in Fig.~\ref{fig:vqvae_results} show that the model preserves the main scene layout and large objects while discarding fine texture. The chosen input size and latent resolution keep the robot-side pipeline lightweight enough for embedded deployment while still producing visually recognizable edge reconstructions.

\boldpar{Semantic mapping}
The edge node fuses reconstructed RGB-D, LiDAR, odometry, and TF through RTAB-Map to build a 3D map and robot trajectory. To extend the map beyond geometry, a fine-tuned YOLO detector is applied to reconstructed images and the detected objects are projected into the map. This allows the \semcom stream to support object-level functions such as semantic search, semantic map inspection, and future go-to-object actions. 

The key point is the information path from compressed tokens to robotics utility: semantic tokens are first decoded into images, detections are then extracted from those reconstructed views, and the resulting object hypotheses are attached to the spatial map and exposed to the operator interface. This makes reconstruction quality meaningful beyond PSNR alone, because the practical question is whether the recovered scene remains informative enough for edge perception and human supervision.

The semantic mapping component demonstrates how reconstructed visual information can be connected to robotics utility. Object detections are associated with depth and pose information, projected into the map frame, and exposed as queryable semantic objects through the dashboard. The present demo focuses on the integration path rather than object detection benchmarking.

\begin{figure*}[t]
    \centering
    \includegraphics[width=\linewidth]{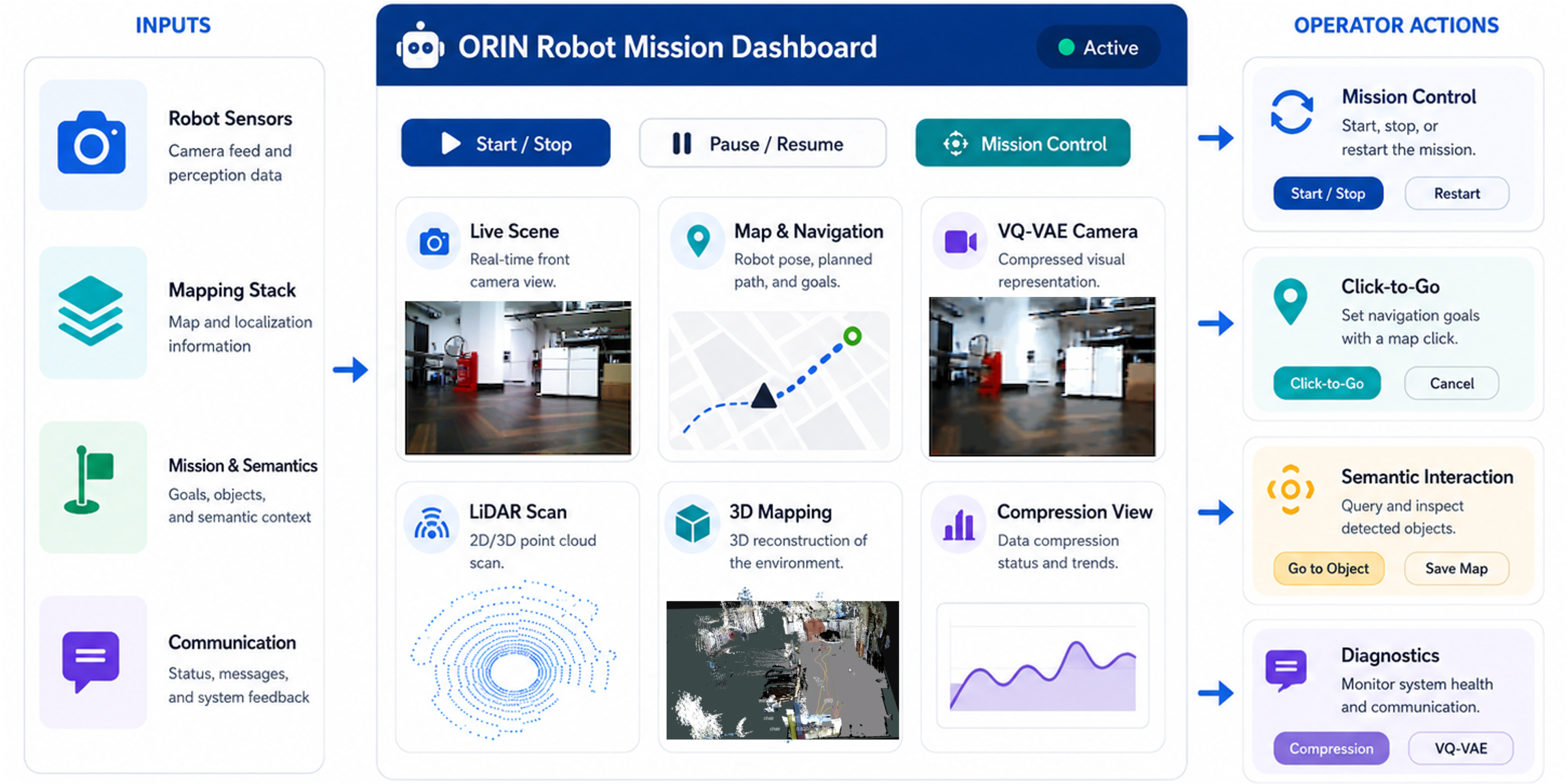}
    \caption{Dashboard functions illustration that combines monitoring, mapping, semantic reconstruction, and mission control.}
    \label{fig:dashboard}
\end{figure*}

\boldpar{Dashboard and integration}
The operator dashboard in Fig.~\ref{fig:dashboard} aggregates the main system views in one place: live camera, LiDAR scan, 2D navigation, 3D mapping, reconstructed semantic camera, and compression status. It also exposes mission actions such as start/stop, pause/resume, and click-to-go navigation. Importantly, dashboard actions are routed through a mission API instead of direct web-to-robot velocity control, so goal requests can be validated against the map and navigation stack before execution. This integration makes the demo useful for studying how communication, perception, and mission control interact under a task-oriented 6G architecture.

Lastly, the dashboard allows viewers to compare raw and reconstructed imagery against map updates and navigation actions within the same time window, making it easier to judge whether semantic compression is merely compact or genuinely useful for robotic decision support.


\section{Task-Oriented Relevance}
\begin{table}[t]
\centering
\caption{Task-oriented observables exposed by the demo.}
\label{tab:task_observables}
\footnotesize
\begin{tabular}{p{0.28\linewidth}p{0.62\linewidth}}
\toprule
\textbf{Observable} & \textbf{Role in the demo} \\
\midrule
Payload reduction & Quantifies the communication load of semantic visual transport. \\
Reconstruction quality & Indicates whether compressed visual information remains interpretable. \\
Semantic map utility & Shows whether reconstructed views can support object-level map interaction. \\
Mission interaction & Exposes whether edge outputs remain actionable through dashboard commands. \\
\bottomrule
\end{tabular}
\vspace{-2mm}
\end{table}

The demo is designed to expose metrics above packet-level QoS. In addition to payload reduction and reconstruction fidelity, the platform makes it possible to observe whether the edge map remains coherent, whether semantic objects stay queryable, whether the operator keeps situational awareness through reconstructed imagery, and whether mission commands derived from edge perception remain valid. These are closer to the application-level outcomes expected in task-oriented 6G systems than throughput or latency alone.

At the networking layer, the setup also provides concrete integration points for a private 5G/Open RAN deployment. The semantic token stream can be treated as a distinct task-relevant flow, dashboard and mission feedback can serve as application-to-network context, and different robot states can imply different communication priorities. For example, navigation through cluttered regions may require lower-latency semantic updates than background mapping. The current implementation therefore acts as a practical bridge between \semcom algorithms and system-level 6G experimentation.

\boldpar{Future work}
The current version focuses on end-to-end integration from robot-side semantic encoding to edge-side mapping and dashboard-assisted mission interaction. Detailed runtime comparisons like wireless latency characterization over the 5G/Open RAN path, energy consumption analysis, and multi-robot task execution and evaluation are left to future work.

\section{Conclusion}

We presented a WIP robot-edge platform for collaborative robotics, integrating visual semantic compression, semantic mapping, and dashboard-based mission control. The current implementation shows how RGB observations can be encoded into compact VQ-VAE token streams on the robot, reconstructed at the edge, and further associated with depth, pose, and 3D mapping information to support object-level semantic interaction. Through the dashboard, the demo makes the full loop visible, from semantic visual transport to map-based object inspection and mission-level control.
The results demonstrate the feasibility of using semantic representations to reduce visual communication load while preserving sufficient information for coarse environment understanding and operator supervision. More broadly, the platform provides a practical basis for studying task-aware 6G robotic networking.

\section*{Acknowledgment}
This work is supported by the 6G-GOALS project under the 6G SNS-JU Horizon program, n.101139232.
\bibliographystyle{IEEEtran} %
\bibliography{IEEEabrv,references} 
\end{document}